\documentclass[letterpaper, 10 pt, conference]{ieeeconf}  

\IEEEoverridecommandlockouts                              

\usepackage{graphicx} 
\usepackage{url}
\usepackage{subcaption}

\title{\LARGE \bf
QLAUN: A Research-Oriented, Robust, Agile, Modular, and Affordable Torque-Controlled Quadruped Robot
}

\author{Mohamad S. Moudallal$^{1}$ and Noel J. Maalouf$^{2}$
\thanks{Presented as an Extended Abstract at the IEEE ICRA@40 Anniversary Conference, Rotterdam, Netherlands, September 2024.}%
\thanks{This work was supported by the PIRF-I0022 grant through the President's Intramural Research Fund at the Lebanese American University.}
\thanks{$^{1}$Mohamad S. Moudallal is the corresponding author and is with the Department of Electrical and Computer Engineering, Lebanese American University, Byblos, Lebanon.
        {\tt\small mohamadsaid.moudallal@lau.edu}}%
\thanks{$^{2}$Noel J. Maalouf is with the Department of Electrical and Computer Engineering, Lebanese American University, Byblos, Lebanon.
        {\tt\small noel.maalouf@lau.edu.lb}}%
}

\begin{document}

\maketitle
\thispagestyle{empty}
\pagestyle{empty}

\begin{abstract}

QLAUN Bot (Quad-Legged Adaptive Unmanned Navigator Robot) is a torque-controlled quadruped robot that is research-oriented, cost-effective, and aimed at achieving simultaneous robustness and agility while being completely 3D-printed. It is a quadruped robot that is aimed at empowering robotics research at universities and research institutes in Lebanon and the MENA region. Using a novel electronics-free leg design strategy, we present a modular robot with interchangeable and easily replaceable legs. The 15 kg robot possesses 12 DoF (Degrees-of-Freedom) with three per leg, each paired with a completely 3D-printed Quasi-Direct Drive (QDD) actuator that consists of a brushless DC motor and a low-ratio gearbox transmission that is connected to a belt transmission system for significantly increasing the torque outputs at the joints. We present legs that have decoupled hip and knee actuators to improve the overall modularity of the robot. QLAUN is almost completely 3D-printed using polylactic acid (PLA) and assembled using off-the-shelf parts to create a robust, agile, and affordable robot for legged robot locomotion research. The legs possess joints with wide ranges of motion, including a continuous hip flexion-extension joint. A compliant foot, printed using TPU-95A is also implemented for alleviating hard impacts and handling terrain uncertainties. This extended abstract aims to introduce QLAUN, a novel platform for robotics research, emphasizing the design concepts and principles that underpin its development to the academic and research communities in the field of robotics.

\end{abstract}

\section{Introduction}

Quadruped robots have existed for some time, with one of the earliest actively actuated models dating back to 1986 with Marc Raibert's quadruped robot \cite{raibert1986legged}. Raibert made significant contributions to the field through his work on leg design, single-leg modeling, and leg control. Over time, quadruped robots have become lighter, faster, and more durable, with improved perception of their surroundings. The development of quadruped robots is a multi-disciplinary process that requires significant resources and effort in areas such as mechanical engineering, locomotion control, perception, and path planning. However, the most critical aspect of the development process is the mechanical design, as it dictates the passive locomotion model of the robot. 

\begin{figure}[htbp]
\centerline{
    \includegraphics[width=\linewidth]{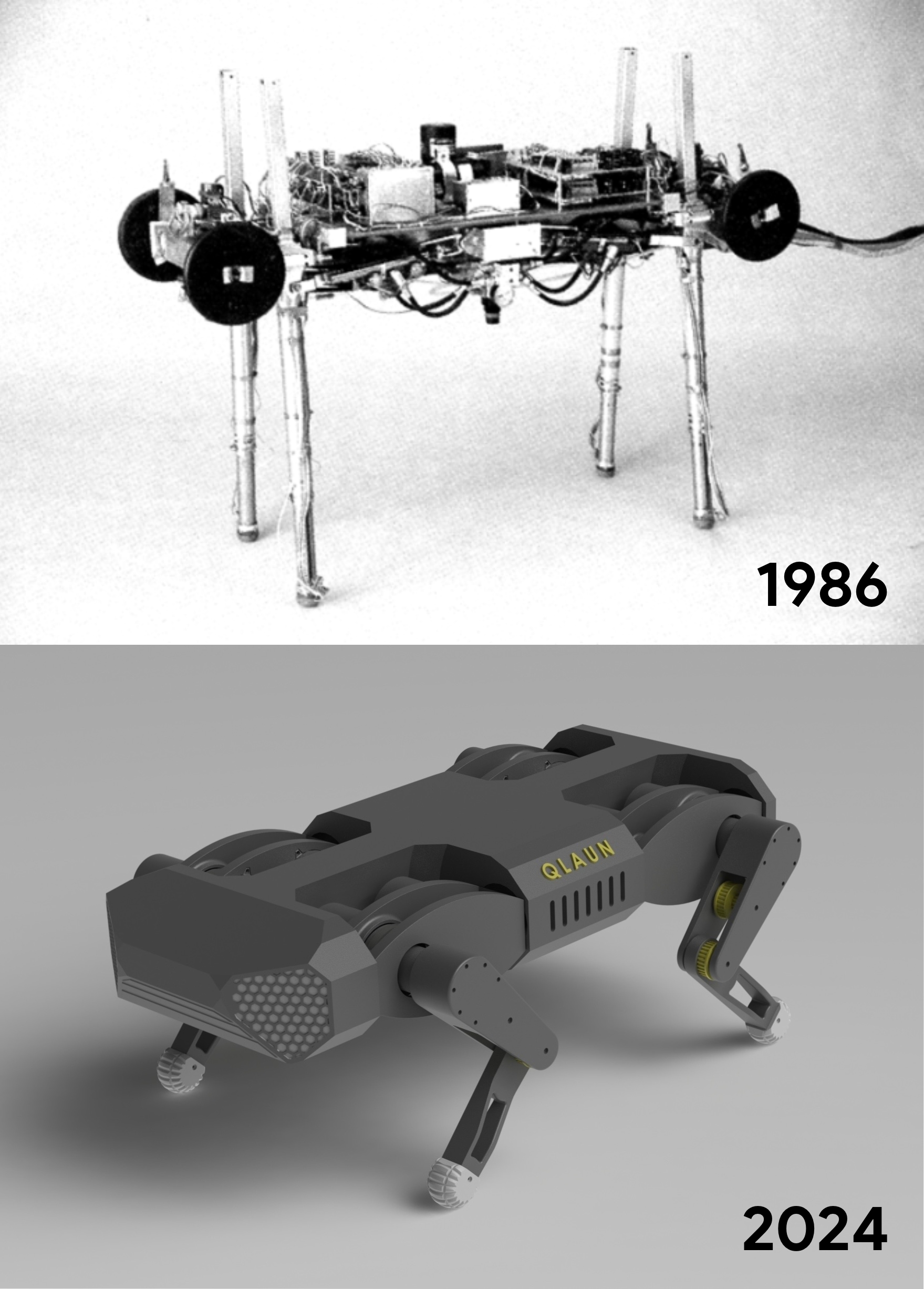}
}
\caption{Visual Comparison between Marc Raibert's Quadruped Robot \cite{raibert1986legged} and QLAUN}
\label{fig1}
\end{figure}

Research on quadruped robotics has extensively explored various leg configurations, each offering distinct advantages for locomotion dynamics and stability. Mainly, three types dominate the field. Prismatic legs are foundational, characterized by their two degrees of freedom (DoF); a revolute joint at the hip and a linear joint connecting the shank to the thigh. This configuration, prominently featured in the pioneering legged robots developed by Marc Raibert \cite{raibert1986legged}, set a precedent for early robotic designs. Articulated legs, that incorporate two DoF at both the hip and knee, represent an evolution in design, providing greater flexibility and control. Extending this concept further, redundant articulated legs introduce an additional joint and link, akin to the anatomical structure of an ankle and foot in quadruped animals \cite{zhong2019analysis}.


Research on legged robotic systems is generally financially challenging for smaller institutions with small to almost no funding. Open-source quadruped robots that are designed for robotics research have been proven to be useful when it comes to investigating and advancing locomotion principles, especially with robots like the robot SOLO from the Open Dynamic Robot Initiative \cite{grimminger2020open}. SOLO is an open-source quadruped robot that is research-oriented, affordable, and very lightweight with a total weight of 2.2 kg, making it very agile.

Most quadruped robots focus mainly on agility, speed, and mobility whereas a low payload-to-weight ratio characterizes them. Kirin, a quasi-direct-drive actuated quadruped robot with prismatic legs, weighs 50 kg and can handle loads up to 125 kg \cite{zhou2022kirin:}. In an analysis done by Y. Zhou et al., the payload-to-weight ratios of the well-known quadruped robots in the industry are presented. All showcased systems feature a payload capacity that is less than 100\%, except for Kirin. Nevertheless, Kirin is a payload-focused robot with prismatic legs augmenting the payload-carrying capacity of the robot, while decreasing the agility of the system dramatically. Kirin has agility and mobility that is nowhere near the ones achieved by agility-oriented quadruped robots such as StarlETH, ANYmal, and Boston Dynamics’ Spot, among others \cite{zhou2022kirin:}.

Systems that accomplish versatility and efficiency at the same time as well as payload-to-weight ratios higher than unity are scarce. There is minimal research on the differences between leg topologies. There is little to no research on leg design alterations between the fore and hind legs with most of them being limited to minor length, angle, or orientation changes \cite{kazama2015development}. Moreover, the field fails to provide a low-cost 3D-printed mechanical robotic quadruped leg that is both robust and agile.

Henceforth, we present QLAUN (Quad-Legged Adaptive Unmanned Navigator), a novel 3D-printed, torque-controlled quadruped robot that is research-oriented, affordable, and modular, aiming to achieve a balance between robustness and agility.

\section{Hardware Overview}

\subsection{Actuator Design}
The main actuator for QLAUN consists of a brushless DC motor (mjbots MJ5208, 330KV, 600W), a 3D-printed 8:1 planetary gearbox to achieve a Quasi-Direct Drive (QDD), and a high-resolution incremental encoder (AMT 103-V, 2048PPR). The actuator is around 85cm in diameter and 65cm tall. The actuator's gearbox possesses helical gears, improving torque output and reducing backlash. QLAUN's actuator is handled and controlled by BLDC drivers that are capable of Field-Oriented Control (ODriveRobotics ODrive Pro). The actuators are operated at 20V and can consume up to 30A from a Li-ion battery. This actuator provides torques of up to 5N.m while limited to 40\% of its full power capacity.

\subsection{Leg Design}
We present QLAUN's leg, called RAPID (Robust \& Affordable Prototyping of an Inertial Driven) Leg, as a novel, completely 3D-printed, and electronics-free design. It is a 3-DOF leg featuring hip abduction-adduction (HAA), hip flexion-extension (HFE), and knee flexion-extension (KFE) joints. The thigh and shank are dimensioned with a 1:1 ratio, as observed in canine mammals \cite{lin2019bionic}. RAPID features decoupled hip (HFE) and knee (KFE) joints by using transmission belts. They both have 4:1 ratios bringing the total ratio for these joints to 32:1, thus augmenting the torque. The electronics-free design follows the modularity principle by allowing researchers to swap out the leg for any custom-made leg using three bolts. The leg is printed using polylactic acid (PLA), assembled with off-the-shelf parts, and features a compliant foot, printed with TPU-95A.

\subsection{System Design}

QLAUN's design philosophy embraces modularity not only in its functional aspects but also in its structural components. The robot's 3D-printed chassis enables a high degree of customization, accommodating everything from specialized sensor arrays to complex manipulator arms. This adaptability makes QLAUN an ideal platform for a broad spectrum of robotics experiments, catering to specific research needs across various domains. Weighing approximately 15 kg and measuring 38 cm tall, 77 cm long, and 38 cm wide, QLAUN combines a compact and manageable size with a durable structure that is conducive to both laboratory and real-world applications. Its versatile design simplifies the process of exploring different leg topologies and sensor integrations, thereby accelerating innovation in robotic mobility and interaction with environments.

\section{Conclusion \& Future Works}

Designed to enhance legged locomotion studies, QLAUN offers a robust, agile, and affordable platform that promotes diverse experimental studies, beginning at the Lebanese American University and extending across the MENA region. QLAUN is set to support advanced research projects, including robust locomotion, SLAM technologies, and terrain analysis. Furthermore, QLAUN will also be available as a ready-made product and as an open-source initiative, providing its designs and software to the global research community to foster innovation and practical applications.

\addtolength{\textheight}{-12cm}   



\bibliographystyle{abbrv}
\nocite{*}
\bibliography{bibliography}

\end{document}